\documentclass[letterpaper, 10pt, journal, twoside]{ieeetran}

\usepackage{cite}
\usepackage{amsmath}
\usepackage{amssymb}
\usepackage{algorithm}
\usepackage{algpseudocode}
\usepackage{xcolor} 
\usepackage{wrapfig} 
\algrenewcommand{\algorithmiccomment}[1]{\hfill \textcolor{gray}{// #1}}
\usepackage{amsmath}
\usepackage{amsfonts}
\usepackage{amssymb}
\usepackage{booktabs}
\usepackage{soul}

\usepackage{cite}
\usepackage{amsmath,amssymb,amsfonts}
\usepackage{graphicx}
\usepackage{textcomp}
\usepackage{xcolor}

\usepackage{subcaption} 
\usepackage{float}
\usepackage{multirow}
\usepackage{pifont}
\usepackage{makecell}

\usepackage{enumitem}
\usepackage{subcaption} 

\usepackage[capitalize]{cleveref}
\crefname{figure}{Fig.}{Figs.}
\crefname{table}{Tab.}{Tabs.}
\crefname{equation}{Eq.}{Eqs.}
\crefname{section}{Sec.}{Secs.}

\usepackage{amsthm}
\theoremstyle{definition}

\usepackage{xspace}
\newcommand{\ie}{\textit{i.e.}\xspace}

\begin{document}

\title{Self-Imitated Diffusion Policy for Efficient and Robust \\ Visual Navigation}

\author{Runhua Zhang$^{1,2}$, Junyi Hou$^{1,2}$, Changxu Cheng$^{1*}$, Qiyi Chen$^1$, Tao Wang$^1$, and Wuyue Zhao$^1$ 
\thanks{Manuscript received: January 6, 2026; Revised: March 8, 2026; Accepted: May 5, 2026. }
\thanks{This paper was recommended for publication by Editor Pascal Vasseur upon evaluation of the Associate Editor and Reviewers comments.}
\thanks{This research was supported by the Key Scientific Research
Program of Hangzhou Municipal Bureau of Science and
Technology grants 2025SZD1A01.}
\thanks{$^1$All authors are with the Uni-Ubi Techonlogy Co., Ltd. Hangzhou, China.}
\thanks{$^2$The 2 \textit{co-first} authors are also with the College of Control Science and Engineering, Zhejiang University, Hangzhou, 310027. e-mail: 22432061@zju.edu.cn}
\thanks{$*$\textit{Corresponding} author: ccx0127@gmail.com}
\thanks{Digital Object Identifier (DOI): see top of this page.}
}

\markboth{IEEE Robotics and Automation Letters. Preprint Version. Accepted May, 2026}%
{Zhang \MakeLowercase{\textit{et al.}}: Self-Imitated Diffusion Policy for Efficient and Robust Visual Navigation}


\maketitle

\begin{abstract}
Diffusion policies (DP) have demonstrated significant potential in visual navigation by capturing diverse multi-modal trajectory distributions.
However, standard imitation learning (IL), which most DP methods rely on for training, often inherits sub-optimality and redundancy from expert demonstrations, thereby necessitating a computationally intensive ``generate-then-filter'' pipeline that relies on auxiliary selectors during inference.
To address these challenges, we propose \textit{Self-Imitated Diffusion Policy (SIDP)}, a novel framework that learns improved planning by selectively imitating a set of trajectories sampled from itself.
Specifically, SIDP introduces a reward-guided self-imitation mechanism that encourages the policy to consistently produce high-quality trajectories efficiently, rather than outputs of inconsistent quality, thereby reducing reliance on extensive sampling and post-filtering.
During training, we employ a reward-driven curriculum as a gated cold-start protector to ensure training stability, and goal-agnostic exploration as a regularizer to preserve the diffusion model’s inherent multi-modal distribution.
Extensive evaluations on a comprehensive simulation benchmark show that SIDP significantly outperforms previous methods, with real-world experiments confirming its effectiveness across multiple robotic platforms.
On Jetson Orin Nano, SIDP delivers a 2.5$\times$ faster inference than the baseline NavDP, \ie, 110ms VS 273ms, enabling efficient real-time deployment.
Project page: https://rhzhang1.github.io/sidp.github.io/.
\end{abstract}


\section{Introduction}
\IEEEPARstart{D}iffusion-based path planning has emerged as a powerful paradigm for visual navigation~\cite{cai_navdp_2025, sridhar_nomad_2024}. By denoising full trajectories, diffusion policies maintain spatiotemporal consistency and capture multi-modal route distributions, making them suitable for complex environments with multiple feasible paths.

\begin{figure}[t]
    \centering
    \includegraphics[width=0.9\columnwidth]{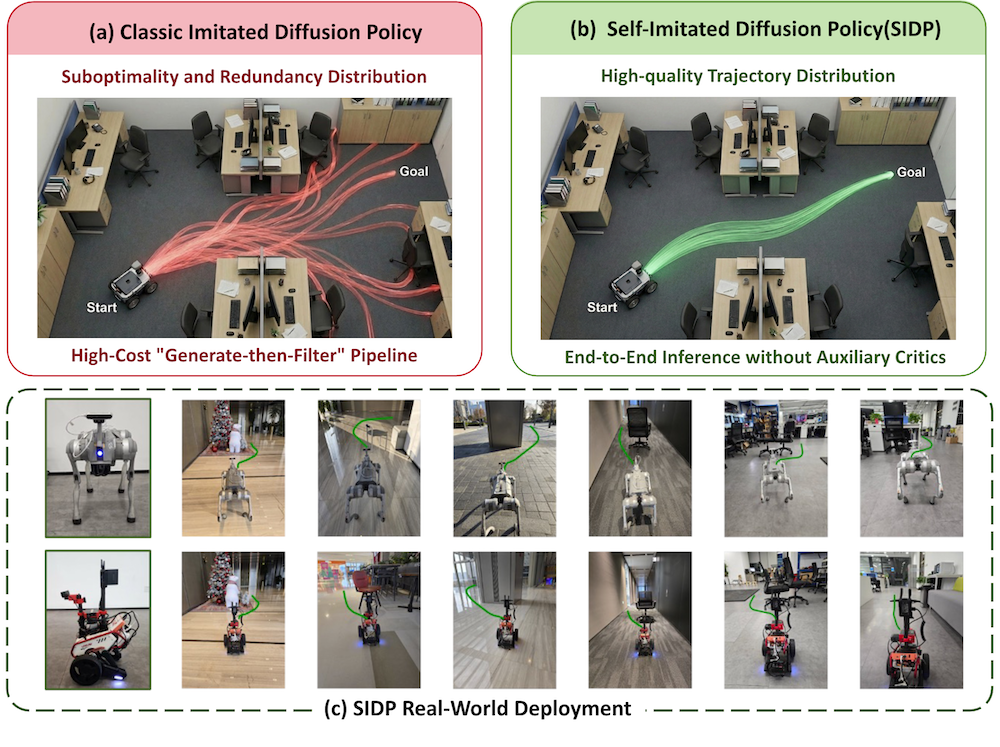}
    \caption{Diffusion policy comparison. 
(a) Imitation yields variable (unsafe/suboptimal) trajectories, requiring dense sampling and filtering. 
(b) SIDP uses reward-guided self-imitation to concentrate the distribution and plan efficiently without selectors. 
(c) Real-world validation on two robotic platforms.}
    \label{fig:intro}
\vspace{-6mm}
\end{figure}

A common training strategy is imitation learning (IL) from expert demonstrations, but IL introduces two bottlenecks.
First, limited dataset coverage weakens robustness under distribution shifts and in novel scenarios~\cite{popov_mitigating_2025, shi_dagger_2025}.
Second, imitating all demonstrated trajectories---including suboptimal ones---increases sampling variance and output inconsistency.
As a result, inference often relies on a ``generate-then-filter'' pipeline with auxiliary trajectory selectors~\cite{cai_navdp_2025,zeng_navidiffusor_2025,bar2025navigation,li_generalized_2025}, which adds latency, especially on resource-constrained devices.
These challenges for imitation-based diffusion policies are illustrated in \cref{fig:intro}(a).

In this work, we propose the \textbf{Self-Imitated Diffusion Policy} (SIDP), a framework designed for efficient and robust path planning in visual navigation.

SIDP employs a reward-guided self-imitation scheme~\cite{oh_self-imitation_2018}, where target trajectories are self-generated rather than sourced from demonstrations. 
By reweighting trajectories based on rewards, the framework prioritizes learning from high-quality experiences.
This paradigm allows the agent to explore edge cases via a `trial-improvement' loop. By rectifying its own low-reward behaviors, the agent develops explicit error-correction, fundamentally addressing the covariate shift inherent in traditional imitation learning.
Beyond improving generalization, SIDP circumvents the complexity of RL-tuned diffusion~\cite{wang2022diffusion,ren2024diffusion} by eliminating Backpropagation Through Time (BPTT) across denoising steps, which is often computationally prohibitive and unstable~\cite{yang2025finetuningdiffusionpoliciesbackpropagation}.
By reformulating optimization into an iterative imitation objective, SIDP achieves stable policy gradients efficiently.
Furthermore, this paradigm enhances navigation \textbf{efficiency} through distributional concentration; suboptimal and redundant trajectories are progressively pruned, preserving inter-modal diversity while promoting intra-modal convergence.
This refinement obviates dense sampling and post-filtering, removing the need for an external selector for a more streamlined inference. 
Consequently, deterministic sampling with fewer denoising steps~\cite{song2020denoising} becomes feasible, further accelerating inference.

Under the SIDP framework, we fortify training with two strategies: goal-agnostic exploration, a stochastic regularizer that preserves multi-modality and prevents mode collapse, and a reward-driven curriculum, which acts as a gated cold-start protector to filter noisy, low-potential trajectorssssies and safeguard the pre-trained policy.

Through extensive experiments, SIDP achieves SOTA performance on the InternVLA-N1 S1 Benchmark~\cite{cai_navdp_2025}, outperforming existing methods in SR and SPL.
While the recent baseline degrades significantly when transitioning from synthetic training scenes to complex, high-fidelity test environments, SIDP effectively mitigates this distribution shift.
Specifically, SIDP surpasses NavDP by $\sim$10\% in success rate on the InternScene-Commercial setting, highlighting its superior generalization and robustness in unseen, unstructured environments.
Furthermore, SIDP boosts inference efficiency to 2.5$\times$ on the Jetson Orin Nano without compromising performance, reducing latency from 273\,ms to 110\,ms.
Successful deployment across two robotic platforms confirms that SIDP provides a practical solution for real-world visual navigation.

The key contributions are summarized as follows:

\begin{itemize}  
    \item We propose SIDP, which enhances planning performance via high-quality self-generated experiences. By fostering distributional concentration, SIDP eliminates dense sampling and post-filtering, streamlining the planning process.
    
    \item We introduce goal-agnostic exploration as a multi-modal regularizer and a reward-driven curriculum as a gated cold-start protector to ensure training stability and convergence.

    \item SIDP achieves SOTA performance in SR and SPL on benchmarks and a 2.5$\times$ speedup on Jetson Orin Nano, demonstrating its real-world robotic practicality.
\end{itemize}

\begin{figure*}[!t]
\centerline{\includegraphics[width=0.80\textwidth]{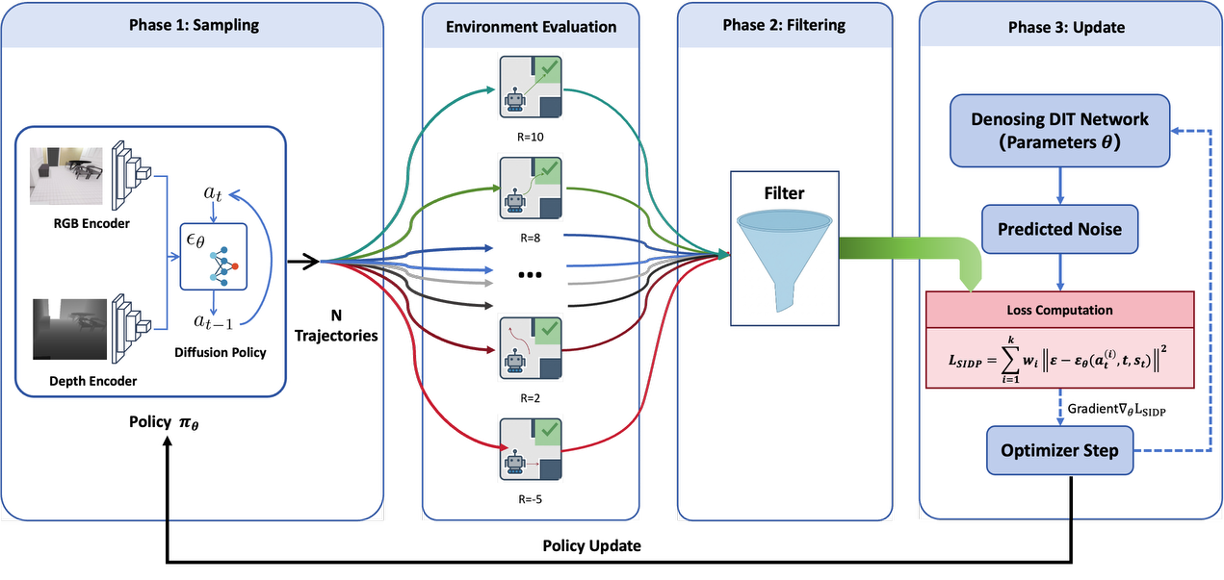}}
\vspace{-3mm}
\caption{
Overview of Self-Imitated Diffusion Policy (SIDP). The framework generates trajectories via $\pi_\theta$, filters them via reward-based sorting, and updates parameters using weighted denoising loss $\mathcal{L}_{\text{SIDP}}$ to align with the optimal distribution.
}
\vspace{-3mm}
\label{frame_pic}
\end{figure*}

\section{RELATED WORK}

\subsection{Path Planning in Visual Navigation}  

Path planning in visual navigation can be categorized into mapping-based and learning-based approaches.
Mapping-based methods \cite{mur-artal_orb-slam2_2017, campos_orb-slam3_2021} rely on visual Simultaneous Localization And Mapping(vSLAM) for map construction and pose estimation, followed by graph-search or sampling-based planners such as Dijkstra, A*, or RRT* to compute feasible paths.
While effective in structured static environments, these methods are constrained by the need for precise mapping and heuristic planning algorithms, reducing robustness in dynamic or perceptually degraded scenarios.

In contrast, learning-based approaches avoid explicit mapping by training policies in simulation or imitating expert demonstrations.
Previous reinforcement learning (RL) methods \cite{wijmans_dd-ppo_2020, ye_auxiliary_2020, desai_auxiliary_2021} use reward signals from environmental priors but incur high computational cost and poor real-world transfer.
Imitation learning (IL) offers better data efficiency \cite{anderson_vision-and-language_2018, fried_speaker-follower_2018, shah_ving_2021, krantz_waypoint_2021, li_infogail_2017, wu_towards_2022}, yet is limited by scarce, low-fidelity expert demonstrations, often suffering from distribution shift. Additionally, IL lacks robust mechanisms to suppress compounding errors, leading to drift in long-horizon navigation. Interactive IL frameworks \cite{ross_reduction_2011, kim_visual_2024} mitigate this via iterative expert dataset refinement.

\subsection{Planning Methods Based on Diffusion Policy}

Diffusion models have recently gained attention in sequential decision-making for capturing multimodal distributions via iterative denoising, outperforming autoregressive \cite{anderson_vision-and-language_2018, fried_speaker-follower_2018, shah_ving_2021, krantz_waypoint_2021} and latent-variable models \cite{li_infogail_2017, wu_towards_2022}. Diffuser \cite{janner_planning_2022} reformulates RL as conditional trajectory generation, while Decision Diffuser \cite{ajay_is_2023} extends this by conditioning on returns and constraints. Subsequent works extend diffusion to long-horizon tasks via hierarchical or compositional designs \cite{liang_skilldiffuser_2024, luo_generative_2025}.

These advances have motivated the application of diffusion policies to path planning.
NoMaD \cite{sridhar_nomad_2024} directly generate trajectories from observations and integrate goal masking for goal-conditioned and undirected exploration. NavDP \cite{cai_navdp_2025} leverages privileged information in simulation for path planning in indoor scenes.
Some works integrate diffusion into classical planning pipelines to improve collision avoidance and planning efficiency \cite{carvalho_motion_2024, yu_ldp_2024}, or employ guided sampling strategies to enforce safety and encourage exploration \cite{zeng_navidiffusor_2025, ren_prior_2025, liao_diffusiondrive_2025, yang_diffusion-es_2024, zheng_diffusion-based_2025, kondo_cgd_2024}.
These approaches demonstrate that diffusion can flexibly model diverse candidate trajectories while incorporating task-specific constraints or guidance.
However, they often require an auxiliary rule-based or learned selector \cite{cai_navdp_2025,li_generalized_2025,bar2025navigation} to assess and filter trajectories generated through diffusion sampling, in order to ensure feasibility and safety.
Such dependencies increase architectural complexity, posing bottlenecks to both computational efficiency and performance during real-world deployment.
These limitations underscore the need for paradigms that intrinsically streamline trajectory generation, enabling robust, end-to-end planning with both theoretical elegance and practical efficiency.

To bridge this gap, we propose the SIDP. SIDP is a generalizable planning framework, whose core is internalizing trajectory selection via self-imitation learning. We focus on visual navigation to demonstrate SIDP’s efficiency, address diffusion-based navigation bottlenecks, and leverage mature evaluation metrics for rigorous validation.

\section{METHOD}
\label{sec:method}

\subsection{Problem Formulation}

We address visual navigation, where an agent generates collision-free optimal trajectories to a specified goal using only egocentric RGB-D observations.
The input state is:
\begin{equation}
s_t = (I_t^{\mathrm{RGB}}, I_t^{\mathrm{D}}, g_t)
\label{eq:inp}
\end{equation}
where the RGB image $I_t^{\mathrm{RGB}} \in \mathbb{R}^{H \times W \times 3}$ and depth map $I_t^{\mathrm{D}} \in \mathbb{R}^{H \times W}$ are camera-observed, and \textit{point goal} $g_t \in \mathbb{R}^3$ is relative to the camera frame.
The policy predicts a sequence of $H$ relative waypoint displacements:
\begin{equation}
a_t = (\Delta p_{t}^1, \dots, \Delta p_{t}^H) \in \mathbb{R}^{H \times 3} = \pi_\theta(s_t).
\end{equation}

To formally evaluate trajectories, we define a quality metric $r(s_t, a_t)$ rewarding safety (collision-free) and efficiency (goal-optimal).
The optimal policy $\pi_\theta$ aims to maximize the expected trajectory quality across the state distribution $\mathcal{D}$:
\begin{equation}
\max_\theta \mathbb{E}_{s_t \sim \mathcal{D},\; a_t \sim \pi_\theta(\cdot \mid s_t)}  r(s_t, a_t).
\label{object1}
\end{equation}

\subsection{Self-Imitated Diffusion Policy}
\label{sec:sidp}

\begin{table}[t]
\centering
\small
\caption{ Components}.
\label{tab:reward_components}
\setlength{\tabcolsep}{10pt}
\renewcommand{\arraystretch}{1.3} 

\begin{tabular}{l l}
\toprule
\textbf{Component} & \textbf{Formulation} \\
\midrule
\multicolumn{2}{l}{\textit{\textbf{Safety}}} \\
Collision & $r_{\text{col}} = -\lambda_{\text{col}} \cdot \mathbb{I}(\text{collision})$ \\
\midrule
\multicolumn{2}{l}{\textit{\textbf{Efficiency}}} \\
Step Cost & $r_{\text{step}} = -\lambda_{\text{step}} \cdot (L_{\text{path}} / d_{\text{init}})$ \\
Progress & $r_{\text{prog}} = \lambda_{\text{prog}} \cdot \Delta d_{\text{geo}}$ \\
Docking & $r_{\text{dock}} = \lambda_{\text{dock}} \cdot \psi(d_t) \cdot \mathbb{I}(d_t < \delta_{\text{fine}})$ \\
\midrule
\textbf{Total} & $r = r_{\text{col}} + r_{\text{step}} + r_{\text{prog}} + r_{\text{dock}}$ \\
\bottomrule
\end{tabular}

\vspace{3pt}
\begin{minipage}{0.95\linewidth}
\footnotesize
\textbf{Settings:} 
$\lambda_{\text{col}}{=}10, \lambda_{\text{step}}{=}0.5, \lambda_{\text{prog}}{=}5, \lambda_{\text{dock}}{=}10$. \\
\textbf{Definitions:} 
$\mathbb{I}(\cdot)$ is the indicator function. $\Delta d_{\text{geo}}$ is geodesic distance reduction. 
$L_{\text{path}}$: trajectory length; $d_{\text{init}}, d_t$: start and terminal distances to goal. \\
\textbf{Docking:} 
$\psi(d) = \exp(-5(d/d_{\text{init}})^2)$ activates within the fine-tuning region $\delta_{\text{fine}}{=}0.5\text{m}$.
\end{minipage}
\vspace{-7mm}
\end{table}

We solve Eq.~\eqref{object1} via distribution matching, \ie, updating policy $\pi_\theta$ to fit the optimal trajectory distribution $p^*(a | s_t)$.
In practice, this is implemented through a simple yet effective self-imitation process.
This is achieved via a simple yet effective self-imitation that bypasses the intractable backpropagation through the iterative denoising chain—a key bottleneck in RL for diffusion policies—by directly updating the policy through distribution matching.

\subsubsection{Distribution Matching}

Maximizing expected reward reduces to matching $\pi_\theta$ with the optimal distribution $p^*$, achieved by minimizing the KL-divergence $D_{\text{KL}}(p^* \| \pi_\theta)$.
This simplifies to maximizing the expected log-likelihood under $p^*$:

\begin{align}
&\arg\min_\theta D_{\text{KL}}(p^* \| \pi_\theta) \nonumber \\
&= \arg\min_\theta \mathbb{E}_{a \sim p^*}\big[ \log p^*(a|s_t) - \log \pi_\theta(a|s_t) \big] \nonumber \\
&= \arg\max_\theta \mathbb{E}_{a \sim p^*}[\log \pi_\theta(a|s_t)].
\label{eq:max_objective}
\end{align}

Since sampling from $p^*$ is intractable, we use $\pi_{\theta}$ as a proposal distribution to generate $N$ candidates $\{a_t^{(i)}\}_{i=1}^N \sim \pi_{\theta}(\cdot | s_t)$, subsequently re-weighted via importance sampling to approximate $p^*$ (\mbox{\cref{frame_pic}}, Phase 1).
The objective thus becomes:
\begin{equation}
\max_\theta \sum_{i=1}^{k} w(a_t^{(i)}) \log \pi_\theta(a_t^{(i)} | s_t).
\label{eq:weighted_ll}
\end{equation}
where $w(a_t^{(i)})$ is the importance weight of the $i$-th candidate.

\subsubsection{Importance Weights}
Directly maximizing Eq. (3) can cause instability. Following REPS~\mbox{\cite{peters2010relative}}, we solve a constrained optimization problem maximizing reward while bounding information loss against the reference $\pi_\theta$:
\begin{equation}
\max_{p} \mathbb{E}_{a \sim p}[r(s_t, a)] \quad \text{s.t.} \quad D_{\text{KL}}(p \parallel \pi_\theta) \leq \epsilon.
\end{equation}
By solving its Lagrangian dual, the optimal distribution $p^*$ is analytically derived as:
\begin{equation}
p^*(a | s_t) \propto \pi_\theta(a | s_t) \exp\left(\frac{r(s_t, a)}{\tau}\right),
\label{eq:optimal_dist}
\end{equation}
where $\tau$ is the Lagrange multiplier governing the trust-region bound $\epsilon$.

To mitigate high variance from Boltzmann distributions, we apply top-$k$ truncation, retaining only the $k$ highest-reward candidates to cover dominant modes of $p^*$, as shown in \mbox{\cref{frame_pic}} Phase 2.
The normalized rewards yield importance weights:
\begin{align}
\boxed{w(a_t^{(i)}) = \frac{\exp(r(s_t, a_t^{(i)})/\tau)}{\sum_{j=1}^{k} \exp(r(s_t, a_t^{(j)})/\tau)}}
\label{3.2}
\end{align}

\subsubsection{Self-Imitated Diffusion Policy Loss}
Inspired by the derivation in DDPM~\cite{ho2020denoising}, maximizing the weighted log-likelihood is reduced to minimizing the reward-weighted denoising loss, which is exactly our SIDP objective:
\begin{equation}
\boxed{\mathcal{L}_{\text{SIDP}} = \sum_{i=1}^{k} w(a_t^{(i)})\, \mathbb{E}_{t_{dn}, \epsilon} \Big[
\bigl\|\epsilon - \epsilon_\theta(a_{t}^{(i), t_{dn}}, t_{dn}, s_t)\bigr\|_2^2 \Big]}
\label{object2}
\end{equation}
where $t_{dn}$ is the diffusion timestep and $\epsilon$ is the noise.

The complete SIDP framework is established (see \cref{alg:sidp}).
By integrating reward-guided self-imitation with diffusion models' representational power, SIDP forms a coherent pipeline for iterative policy refinement.

This ``trial-improvement'' loop endows the policy with explicit error-correction. 
Unlike traditional imitation learning (IL), limited by narrow expert distributions and $\mathcal{O}(T^2)$ compounding errors \mbox{\cite{ross2011reduction}}, SIDP actively explores the state manifold. 
By self-imitating high-reward experiences, SIDP learns corrective mappings for deviated states, fundamentally breaking the error accumulation cycle and enhancing robustness under environmental disturbances.
This mechanism resists execution noise and ensures robust long-horizon navigation.

\begin{algorithm}[h] 
   \caption{Self-Imitated Diffusion Policy (SIDP)}
   \label{alg:sidp}
   \begin{algorithmic}[1]
   \State \textbf{Input:} Policy network \( \pi_\theta \), candidate count \( N \), truncation \( k \), temperature \( \tau \)
   \State \textbf{Initialize:} Environment $\mathcal{E}$
   \While{NOT converged}
       \State \textcolor{gray}{\textit{// Phase 1: Online Sampling}}
       \State Observe current state \( s_t \) from environment $\mathcal{E}$
       \State Generate \( N \) candidate actions \(\mathcal{A}_t\) using \( \pi_\theta \).
       
       \State \textcolor{gray}{\textit{// Environment: Reward Evaluation}}
       \For {each candidate action \( a_t^{(i)} \in \mathcal{A}_t \)}
           \State Evaluate action quality: \( R_i = r(s_t, a_t^{(i)}) \) 
       \EndFor
       
       \State \textcolor{gray}{\textit{// Phase 2: Filtering -- Importance weighting}}
       \State Select top-\( k \) actions with highest \( R_i \): \( \mathcal{A}_t^{\text{top-}k} \)
       \State Compute normalized importance weights (Eq.~\ref{3.2}).
       
       \State \textcolor{gray}{\textit{// Phase 3: Parameter Update (On-Policy)}}
        \State Sample diffusion step \( t \) and noise \( \epsilon \sim \mathcal{N}(0, I) \)
        \State Compute weighted denoising loss (refer to Eq.~\ref{object2}).
        \State Update policy parameters using \(\nabla_\theta \mathcal{L}_{\text{SIDP}}\)
   \EndWhile
   \State \textbf{Output:} Optimized policy \( \pi_\theta \)
\end{algorithmic}

\end{algorithm}

\subsection{Complementary Learning Strategies}
\label{sec:practical_strategies}

To further augment the capabilities of the SIDP framework,
we devise two synergistic strategies: goal-agnostic exploration and reward-driven curriculum learning.

\subsubsection{Goal-agnostic Exploration}
We augment training with a \textit{goal-agnostic} setting to activate exploration \mbox{\cite{sridhar_nomad_2024}} and regularize navigation learning.
At each reset, we sample auxiliary goals within $-60 \sim 60^\circ$ and 3$\sim$5\,m.
The current policy $\pi_\theta$ then generates feasible secure paths toward these goals.

During training, the point goal $g_t$ in \cref{eq:inp} is replaced by a goal-agnostic embedding, with importance weights in \cref{eq:weighted_ll} set to uniform. 
This strategy facilitates exploration by decoupling policy behaviors from specific objectives.

This acts as spatial regularization, preventing the diffusion policy from collapsing onto narrow goal-conditioned paths. By preserving trajectory diversity while encouraging generalized geometric representations, it improves robustness to environmental variations and cross-scene generalization.

\subsubsection{Reward-Driven Curriculum Learning}

To ensure stability and address cold-start challenges, we dynamically regulate the scenario distribution. This balances two trade-offs: (i) filtering "unlearnable" complex scenarios to suppress gradient noise; and (ii) avoiding trivial ones where vanishing or non-informative gradients impair the policy's ability to generalize and master diverse behaviors.

To operationalize this, we evaluate each scenario $s_i$ using two trajectory-derived metrics: maximum reward $R_{\text{max}}(s_i)$ and reward range $R_{\text{range}}(s_i)$. A scenario is admitted if:
\begin{equation}
    R_{\text{max}}(s_i) \geq \tau_{\text{max}} \quad \land \quad R_{\text{range}}(s_i) \geq \tau_{\text{range}}.
\end{equation}

Specifically, $\tau_{\text{max}}$ ensures feasibility by filtering divergent noise from complex scenarios, shielding the policy from cold-start instability. Simultaneously, $\tau_{\text{range}}$ maintains gradient discriminability by excluding trivial, flat-reward environments. By regulating these dynamics, our mechanism enhances training stability and convergence, improving success rate and path efficiency.

\subsection{Efficient Inference of SIDP}

Benefiting from the concentrated and robust trajectory distributions, SIDP bypasses the costly "generate-then-filter" paradigm. Unlike previous planners~\cite{cai_navdp_2025,sridhar_nomad_2024} that rely on DDPM’s stochasticity to compensate for less refined policies, our model’s high-precision distribution allows for deterministic sampling via DDIM~\cite{song2020denoising}. This transition significantly reduces inference steps and eliminates the need for auxiliary selectors, streamlining the entire planning pipeline.

\begin{table*}[t]
\centering
\small 
\caption{Quantitative comparison on InternVLA-N1 S1. We report SR (\%) and SPL (\%), with mSR and mSPL as mean metrics across scenes. For robustness, Relative Success Retention (RSR) is calculated against the \textit{ClutteredEnv-Easy} baseline; mRSR denotes the mean retention across three domain shifts.}
\label{tab:benchmark}
\setlength{\tabcolsep}{3.5pt} 
\begin{tabular}{l ccc ccc cc ccc ccc} 
\toprule
\multirow{3.5}{*}{\textbf{Method}} &
\multicolumn{6}{c}{\textbf{InternScenes}} &
\multicolumn{5}{c}{\textbf{ClutteredEnv}} &
\multirow{3.5}{*}{\textbf{mSR}($\uparrow$)} &
\multirow{3.5}{*}{\textbf{mSPL}($\uparrow$)} &
\multirow{3.5}{*}{\textbf{mRSR}($\uparrow$)} \\
\cmidrule(lr){2-7} \cmidrule(lr){8-12}
& \multicolumn{3}{c}{Commercial} & \multicolumn{3}{c}{Home}
& \multicolumn{2}{c}{Easy (Base)} & \multicolumn{3}{c}{Hard} & & & \\
\cmidrule(lr){2-4} \cmidrule(lr){5-7} \cmidrule(lr){8-9} \cmidrule(lr){10-12}
& SR & SPL & RSR($\uparrow$) & SR & SPL & RSR($\uparrow$) & SR & SPL & SR & SPL & RSR($\uparrow$) & & & \\
\midrule
iPlanner & 53.02 & 51.44 & 59.04 & 39.16 & 37.27 & 43.61 & 89.80 & 88.70 & 80.69 & 79.28 & 89.86 & 59.14 & 57.57 & 64.17 \\
ViPlanner & 64.43 & 62.90 & 78.50 & 43.61 & 42.07 & 53.13 & 82.08 & 81.98 & 67.24 & 67.07 & 81.92 & 60.90 & 59.83 & 71.18 \\
NavDP & 71.25 & 68.89 & 76.31 & 57.38 & 55.08 & 61.45 & 93.37 & \textbf{91.44} & 88.71 & 86.31 & 95.01 & 73.22 & 70.95 & 77.59 \\
\textbf{SIDP (Ours)} & \textbf{81.19} & \textbf{73.36} & \textbf{86.04} & \textbf{63.17} & \textbf{56.48} & \textbf{66.95} & \textbf{94.36} & 89.86 & \textbf{91.58} & \textbf{86.78} & \textbf{97.05} & \textbf{79.11} & \textbf{72.72} & \textbf{83.35} \\
\bottomrule
\end{tabular}
\vspace{-4mm}
\end{table*} 

\section{Experiments}
\label{sec:Experiments}

\subsection{Evaluation Setup}

\subsubsection{Benchmarks}

For evaluation, we conducted experiments on the \textit{InternVLA-N1 S1} benchmark \mbox{\cite{cai_navdp_2025}} using the high-fidelity Isaac Sim simulator.
This benchmark targets real-time closed-loop navigation, requiring continuous, multi-step goal reaching. 
It comprises 60 diverse scenes: 40 from \textit{InternScenes} (home and commercial) and 20 from \textit{ClutteredEnv} (10 easy, 10 hard). 
For each scene, 100 start-goal pairs were randomly sampled in unoccupied spaces, with initial orientations pre-calculated to ensure collision-free initialization.

For ablation, we curated a representative subset from \textit{InternData-N1}~\cite{cai_navdp_2025}, covering 5 scenes and 500 total trials.
Unlike the real-time \textit{InternVLA-N1 S1}, this setup targets one-shot trajectory generation, evaluating single-inference path planning.
To ensure consistency, we utilized fixed starting positions with randomly sampled goals for each scene.

\subsubsection{Evaluation Metrics}

To facilitate a comprehensive and rigorous evaluation, we report the following standard metrics for the main navigation task:

\begin{itemize}
    \item \textbf{Success Rate (SR, \%):} The percentage of episodes where the agent successfully reaches within 0.5\,m of the target without any collision.
    \item \textbf{Success Weighted by Path Length (SPL, \%):} Evaluates success and efficiency:
    \begin{equation}
        \text{SPL} = \frac{1}{N} \sum_{i=1}^N S_i \cdot \frac{L_i}{\max(P_i, L_i)},
    \end{equation}
    where $S_i \in \{0, 1\}$ indicates success, $L_i$ is the geodesic shortest distance, and $P_i$ is the agent's actual path length.
\end{itemize}

We introduce three additional metrics to analyze safety and exploration quality within our ablation benchmark:
\begin{itemize}
    \item \textbf{Collision Rate (CR, \%):} The ratio of episodes with at least one collision, indicating navigation safety.
    \item \textbf{Distance To Goal (DTG, m):} The average Euclidean distance between the agent's final position and the goal.
    \item \textbf{Exploration Area (EA, m$^2$):} The non-redundant union of $16$ sampled trajectories (each modeled as a $0.2\text{m}$-wide swept area), quantifying exploration diversity.

\end{itemize}

Finally, to evaluate robustness—maintaining success under perceptual degradation and environmental variations. We introduce:

\begin{itemize}
    \item \textbf{Relative Success Retention (RSR, \%):} Inspired by ImageNet-C~\cite{hendrycks2019benchmarking}, RSR isolates degradation to quantify structural stability:
    \begin{equation}
        \text{RSR} = \frac{\text{SR}_{\text{perturbed}}}{\text{SR}_{\text{clean}}} \times 100\%,
    \end{equation}
    where $\text{SR}_{\text{perturbed}}$ and $\text{SR}_{\text{clean}}$ success rates in the simple clean baseline and under visual perturbations across three complex environments, respectively.
\end{itemize}

\subsection{Implementation Details}
\subsubsection{Training Scenario Construction}

To enable efficient self-imitation without physics simulation overhead, we construct an interactive environment based on the \textit{InternData-N1} dataset \mbox{\cite{cai_navdp_2025}}, sharing the exact visual data with the baseline NavDP.

Unlike NavDP, which is restricted to static expert demonstrations, SIDP eliminates the reliance on expert labels during iterative refinement. Leveraging this, we reuse initial visual observations (RGB-D inputs from expert trajectories) but dynamically generate novel goals of varying difficulty using the scene's global Euclidean Signed Distance Field (ESDF).

These composed start-goal pairs share the baseline's visual space but lack expert labels, enabling self-exploration rather than external data augmentation. Instead of supervised labels, collision detection is approximated by querying ESDF values along self-predicted paths. This allows the agent to evaluate trajectory safety against complex geometries and learn purely from environmental rewards, as detailed in \mbox{\cref{tab:reward_components}}.

\subsubsection{Network Architecture and Training Details}

We initialize our policy with the diffusion-based NavDP~\cite{cai_navdp_2025}, omitting the auxiliary critic network. Optimization uses AdamW with a $2 \times 10^{-5}$ learning rate. We employ a 128 effective batch size ($64 \times 2$) across two NVIDIA A100 GPUs, with training taking approximately 35 hours.

\subsection{Comparisons with State-of-the-art Methods}
We compare our method with several learning-based approaches: ViPlanner~\cite{roth2024viplanner} and iPlanner~\cite{yang2023iplanner}, which combine learned perception with analytical planners, and NavDP~\cite{cai_navdp_2025}. As our primary baseline, NavDP shares our diffusion-based architecture but is trained via pure imitation learning, enabling a direct assessment of self-imitation's impact on robustness and efficiency.

\subsubsection{Navigation Performance Analysis}

Quantitative results are summarized in Table~\ref{tab:benchmark}. SIDP achieves the highest SR across all scenarios and superior overall performance (mSR and mSPL). It also attains the highest mRSR, indicating enhanced robustness under scene variations. Notably, SIDP shows the largest gains in the challenging \textit{InternScenes} environments. In \textit{Commercial} and \textit{Home} scenarios, SIDP improves SR by 9.94\% and 5.79\% over NavDP, while boosting RSR by 9.73\% and 5.50\%, respectively. This RSR improvement underscores SIDP’s superior robustness when navigating complex, unseen scene variations.

\subsubsection{Computational Efficiency and Ablation Analysis}
 We evaluate inference latency on the NVIDIA Jetson Orin Nano (8GB RAM)—the same edge platform used in our real-world experiments—to disentangle efficiency gains (\cref{tab:ddim_time_compare}).

\textbf{Architectural Efficiency}: 
While SIDP shares an identical neural backbone with NavDP, it internalizes the selection process into its weights, entirely eliminating the auxiliary trajectory selection module. Under the identical 10-step DDPM configuration (\cref{tab:ddim_time_compare}), SIDP achieves a 2.07$\times$ speedup (132,ms vs. 273,ms) and higher SR (0.670 vs. 0.549). 

\textbf{Robustness to Step Reduction}: SIDP exhibits superior performance retention during aggressive step reduction. At 3 steps, NavDP’s SR collapses to 0.275, while SIDP maintains a robust 0.655. This stability results from SIDP’s concentrated policy distribution, which minimizes action-space variance and mitigates discretization errors in low-step DDIM sampling.
The 5-step DDIM configuration offers the optimal balance, yielding 0.674 SR with a 2.48$\times$ speedup, confirming that SIDP enables high-efficiency reactive navigation.

\begin{table}
    \centering
    \caption{Inference time and Success Rate (SR) comparison between different models and scheduler.}
    \label{tab:complete_time_comparison}
    \begin{tabular}{lcccc}
        \toprule
        Method & Scheduler & Denoising Steps & Time (ms) & SR ($\uparrow$) \\ 
        \midrule
        NavDP & DDPM & 10 & 273 ($1\times$) & 0.549 \\
        NavDP & DDIM & 10 & 271 ($1.01\times$) & 0.548 \\
        NavDP & DDIM & 5  & 142 ($1.92\times$) & 0.412 \\
        NavDP & DDIM & 3  & 106 ($2.58\times$) & 0.275 \\
        \midrule
        SIDP  & DDPM & 10 & 132 ($2.07\times$) & 0.670 \\
        SIDP  & DDIM & 10 & 131 ($2.08\times$) & 0.670 \\
        SIDP  & DDIM & 5  & 110 ($2.48\times$) & \textbf{0.674} \\
        SIDP  & DDIM & 3  & \textbf{99} ($2.76\times$) & 0.655 \\ 
        \bottomrule
    \end{tabular}
    \label{tab:ddim_time_compare}
\vspace{-4mm}
\end{table}

\subsubsection{Robustness Analysis}We evaluate SIDP's robustness across two critical dimensions: sensory degradations and environmental complexities.

\begin{figure}
    \centering
    \includegraphics[width=1.0\linewidth]{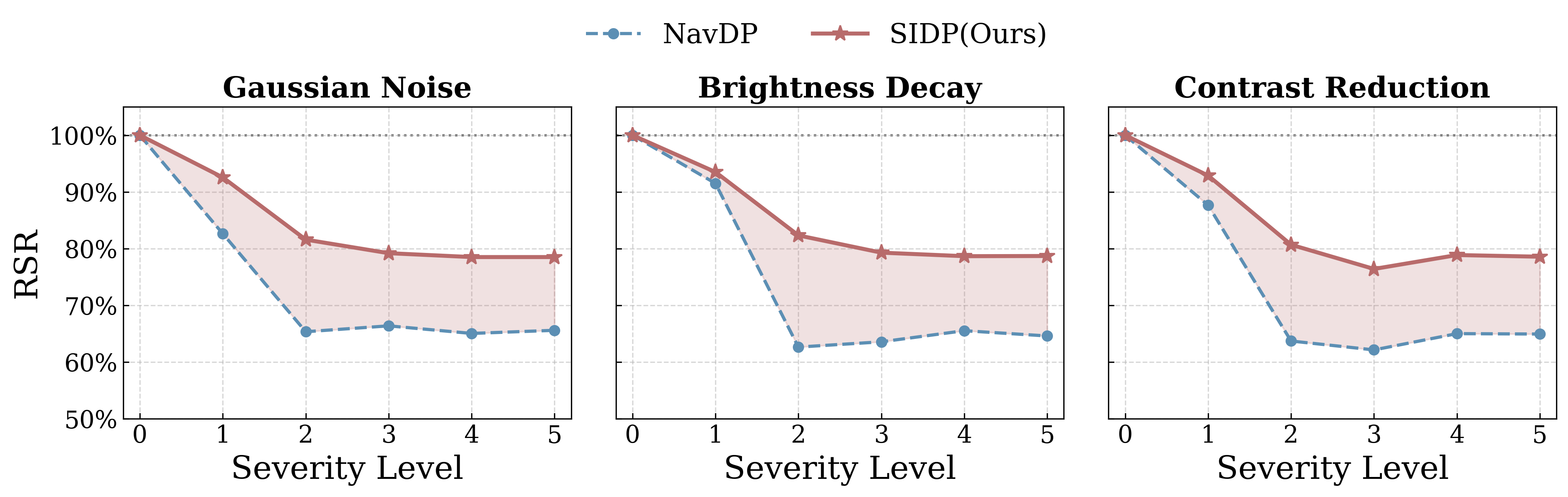}
    \caption{Robustness to corruptions. RSR is normalized to clean conditions; shaded regions show SIDP's performance margin over baseline across severity levels 0--5.}
    \label{fig:robustness}
\vspace{-6mm}
\end{figure}

\mbox{Fig.~\ref{fig:robustness}} compares SIDP with the NavDP baseline under visual corruptions. While NavDP's performance drops below 65\% RSR (severity level 5), SIDP maintains $\sim$80\% RSR, confirming that our self-imitation effectively recovers robust actions from degraded observations

Table~\mbox{\ref{tab:benchmark}} highlights SIDP's resilience to spatial clutter and domain shifts. Transitioning from ClutteredEnv-Easy to Hard, SIDP's Success Rate (SR) degrades by only 2.9\%, outperforming NavDP's 5.0\% drop. Furthermore, in the highly challenging InternScenes-Home domain where traditional methods (e.g., iPlanner~\cite{yang2023iplanner}) collapse (SR $<$ 40\%), SIDP establishes a high performance floor of 63.17\% SR. These results validate SIDP's superior robustness for real-world deployment.

\subsection{Ablation Studies and Analysis}

To evaluate key components, we conduct ablation studies based on the strategies in Section~\ref{sec:method}: 
(1) Reward-Guided Self-Imitation: We investigate its role (\cref{sec:sidp}) in enhancing training stability and accelerating policy convergence; 
(2) Goal-agnostic Exploration: We assess how this strategy (\cref{sec:practical_strategies}) acts as spatial regularization to ensure diffusion output diversity and prevent path collapse; 
(3) Reward-Driven Curriculum Learning: We validate this scheme (\cref{sec:practical_strategies}) as a \textit{gated cold-start mechanism}, filtering unlearnable scenarios via $R_{\text{max}}$ and $R_{\text{range}}$ to guarantee initial stability; and 
(4) Comparison with DAgger~\cite{shi_dagger_2025}: Demonstrating SIDP's advantages over expert imitation on augmented dataset.

\subsubsection{Reward-Guided Self-Imitation}
\label{sec:ablation}
\begin{figure}
    \centering
    \includegraphics[width=0.8\linewidth]{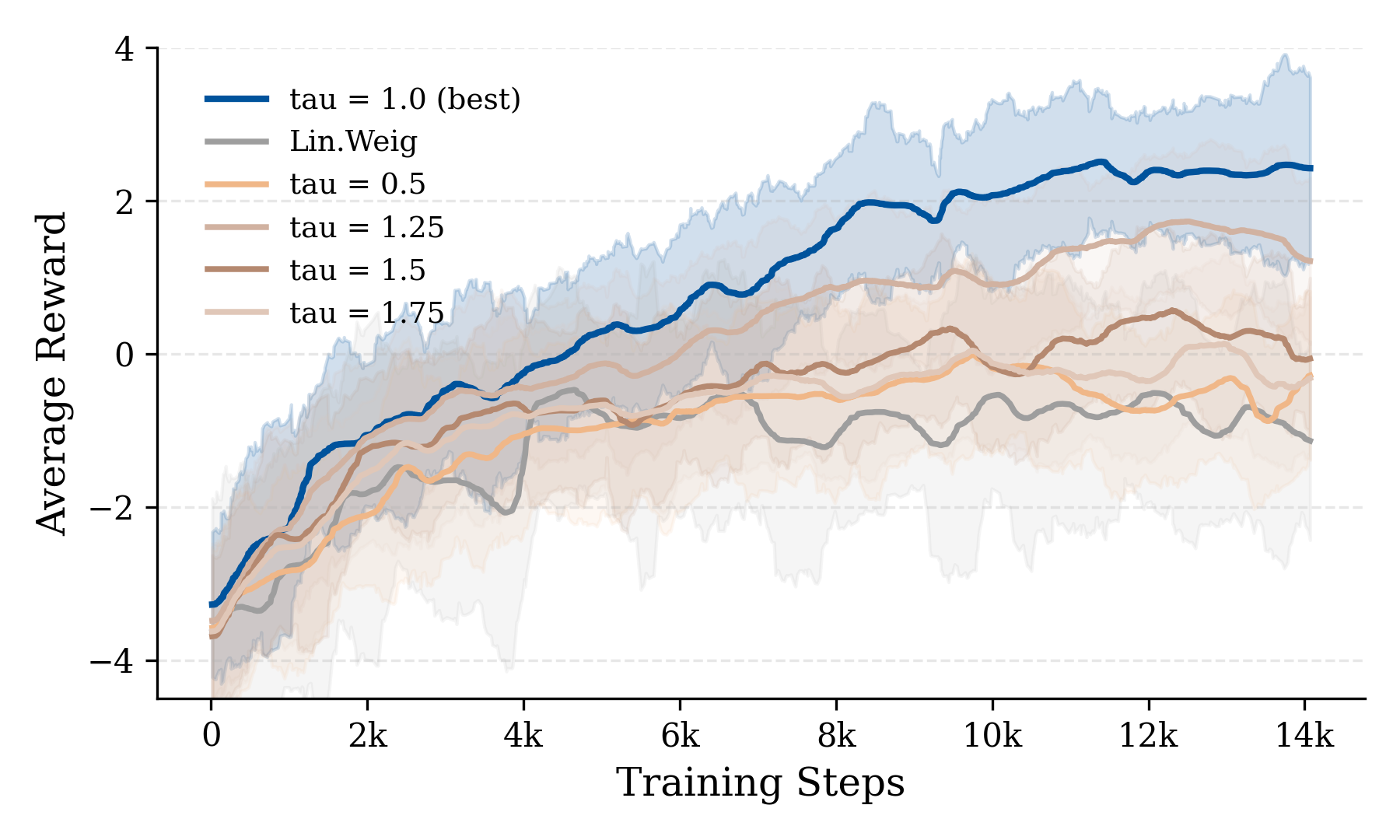}
    \caption{
    SIDP learning curves across temperature coefficients. Curves are Gaussian-smoothed ($\sigma=10$); shaded areas indicate rolling-window standard deviation.
    }
    \label{loss_fig}
\vspace{-4mm}
\end{figure}

\begin{figure}
    \centering
    \includegraphics[width=0.80\linewidth]{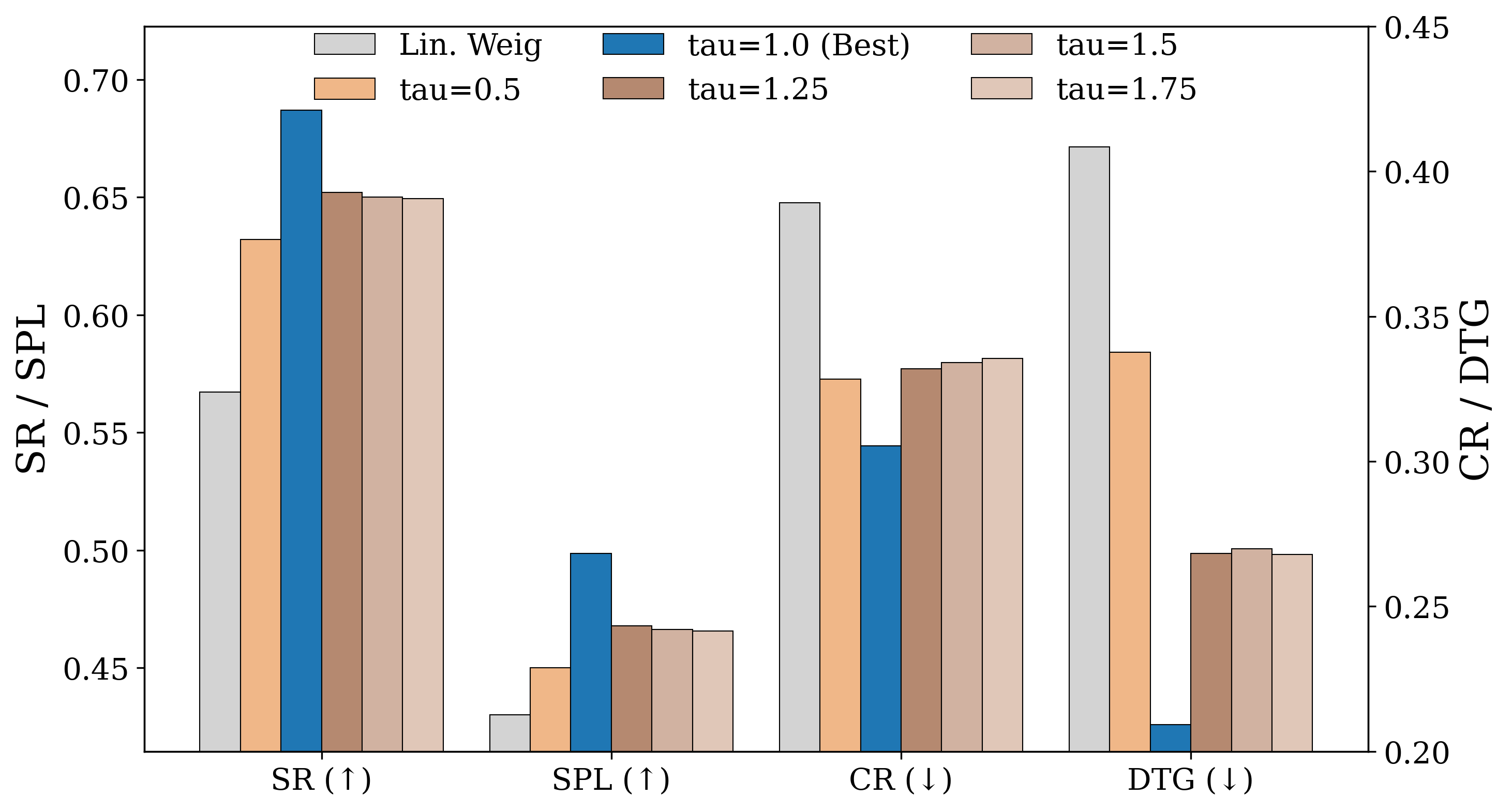}
    \caption{Ablation study of the reward-guided self-imitation mechanism and the Softmax temperature coefficient $\tau$.}
    \vspace{-6mm}
    \label{fig:temp_ablation}
\end{figure}

We also evaluate the temperature coefficient $\tau$, modulating selection sharpness. In \cref{loss_fig,fig:temp_ablation}, $\tau=1.0$ is optimal. Extreme values are detrimental: low $\tau$ causes instability via over-fitting, while high $\tau$ flattens the action distribution.

\subsubsection{Goal-agnostic Exploration}
Fig. 6 shows that goal-agnostic objectives are pivotal for navigation performance and generative diversity. In point-goal tasks, superior SR and SPL indicate effective spatial regularization; by decoupling behaviors from specific targets, the agent learns generalized geometric representations—like corridors and boundaries—avoiding over-fitting to narrow paths.

Crucially, higher EA in exploration tasks provides quantitative evidence of the policy’s multi-modal distribution. Unlike deterministic baselines prone to mode collapse, our approach prevents convergence onto a singular trajectory. By maintaining path diversity and geometric universality, this mechanism ensures a broad distribution of candidate trajectories, essential for cross-scene generalization in unseen environments.

\begin{figure}[t]
    \centering
    \includegraphics[width=1.0\linewidth]{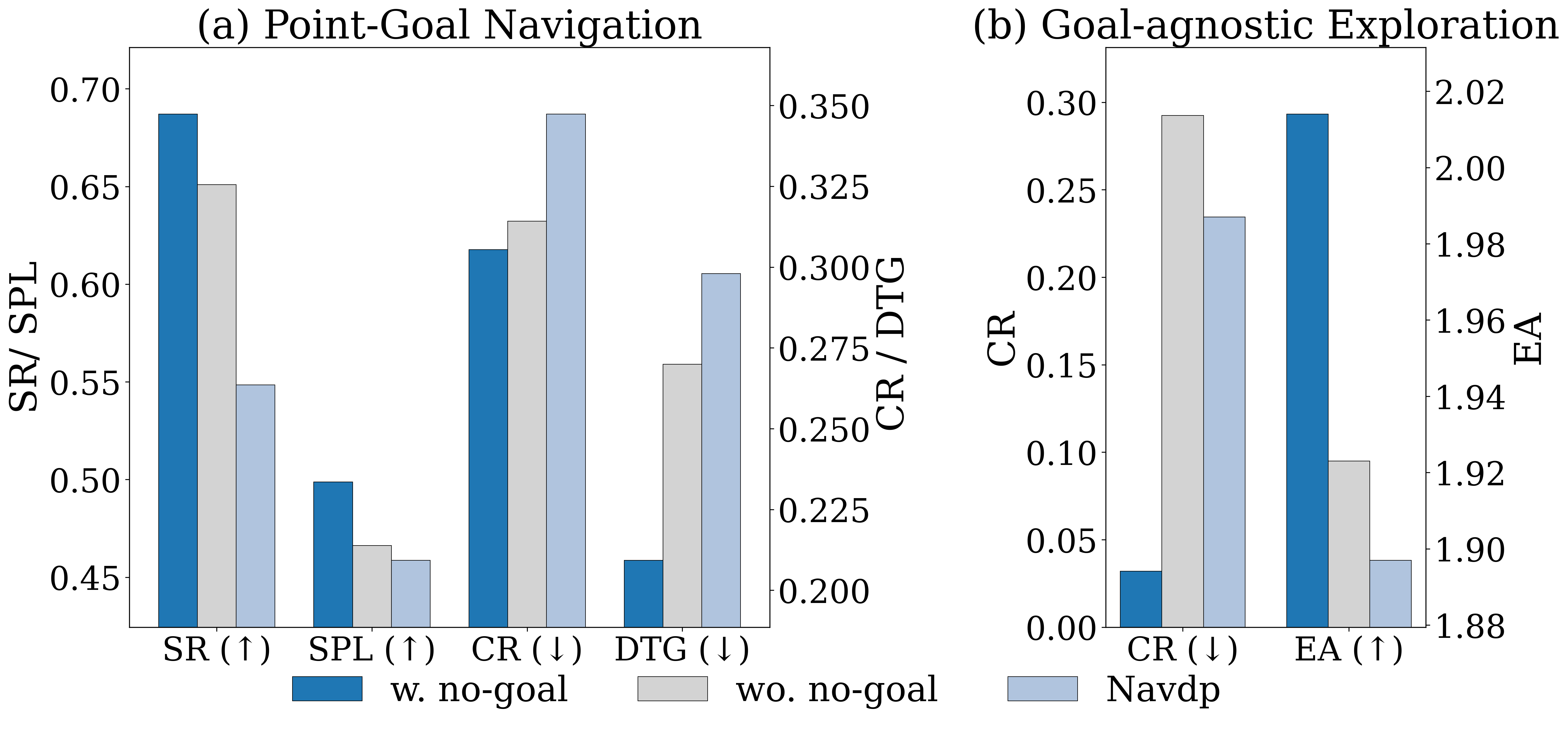}
    \caption{Ablation study of goal-agnostic training under different evaluation settings.}
    \label{fig:nogoal_ablation}
\vspace{-4mm}
\end{figure}

\subsubsection{Reward-Driven Curriculum Learning}
\begin{figure}[t]
    \centering
    \begin{subfigure}[b]{0.85\linewidth}
        \centering
        \includegraphics[width=\linewidth]{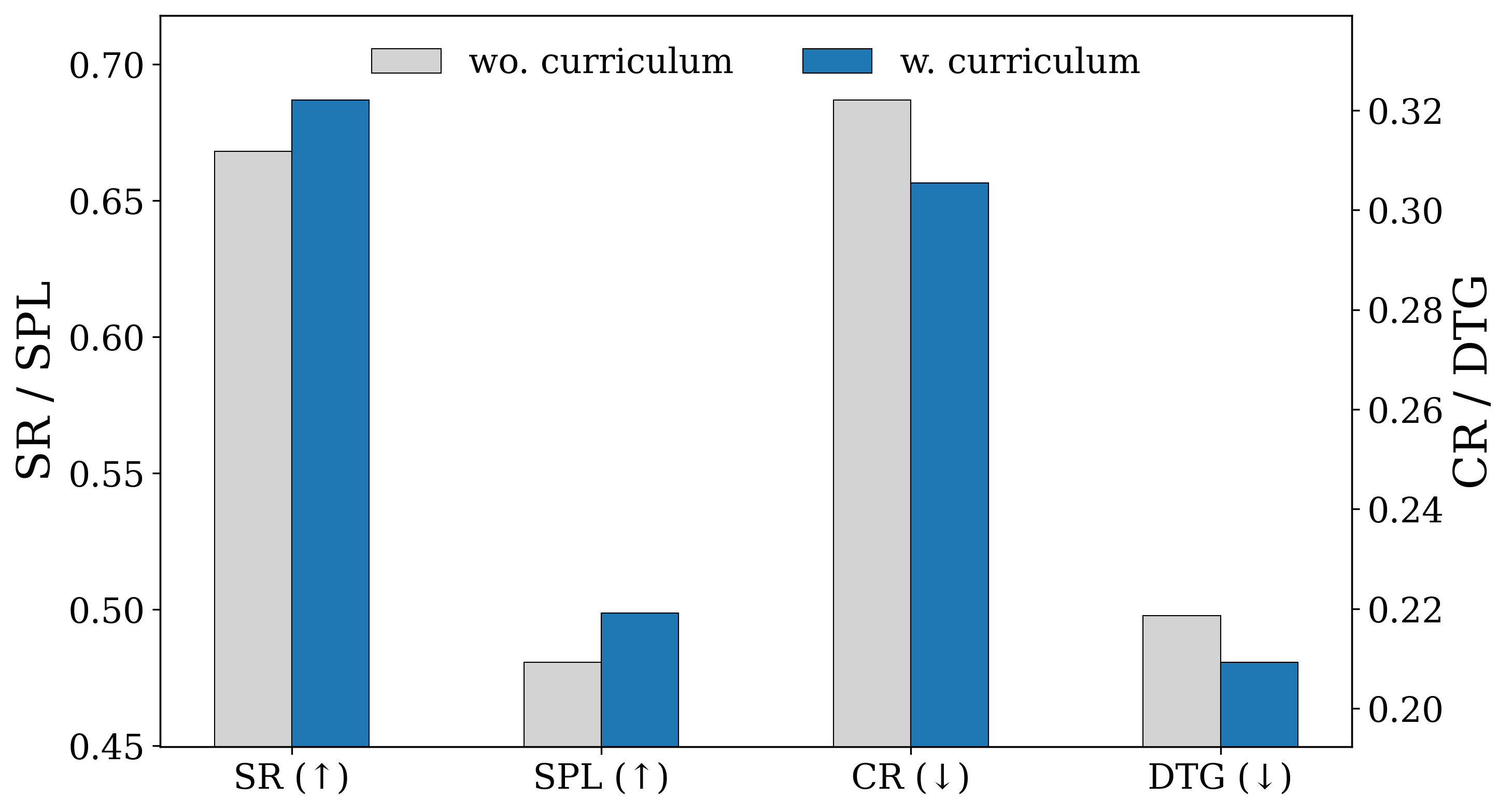}
        \caption{Reward-driven ablation.}
        \label{fig:curriculum_ablation_a}
    \end{subfigure}
    
    \vspace{3mm} 

    \begin{subfigure}[b]{0.85\linewidth}
        \centering
        \includegraphics[width=\linewidth]{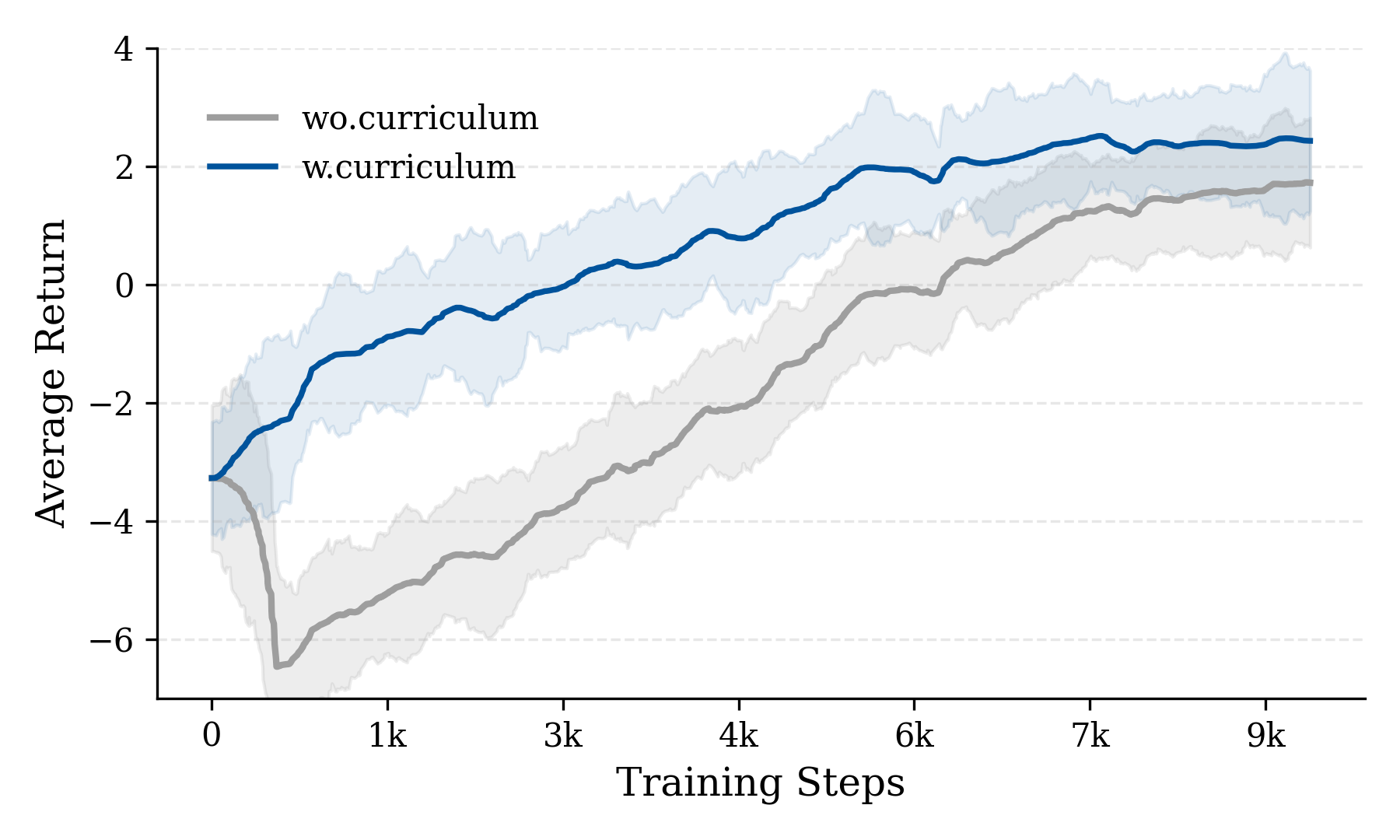}
        \caption{Cold-start analysis.}
        \label{fig:coldstarter_b}
    \end{subfigure}

    \caption{Performance analysis: (a) Curriculum ablation; (b) Cold-start effect.
    }
    \label{fig:curriculum_ablation}
    
    \vspace{-4mm} 
\end{figure}

We evaluate the curriculum's impact on training and performance. As in Fig.~\mbox{\ref{fig:curriculum_ablation}}(a), SIDP achieves superior SR and lower CR, confirming that \mbox{$\tau_{\text{max}}$} suppresses destabilizing gradient noise from failed trajectories. Gains in SPL and reduced DTG demonstrate that \mbox{$R_{\text{range}}$} ensures the gradient discriminability needed for path efficiency. Notably, curves in Fig.~\mbox{\ref{fig:curriculum_ablation}}(b) show our mechanism mitigates cold-start instability, ensuring monotonic convergence versus the baseline's early-stage collapse.

\begin{figure}[t] 
    \centering
\
    \includegraphics[width=0.9\linewidth]{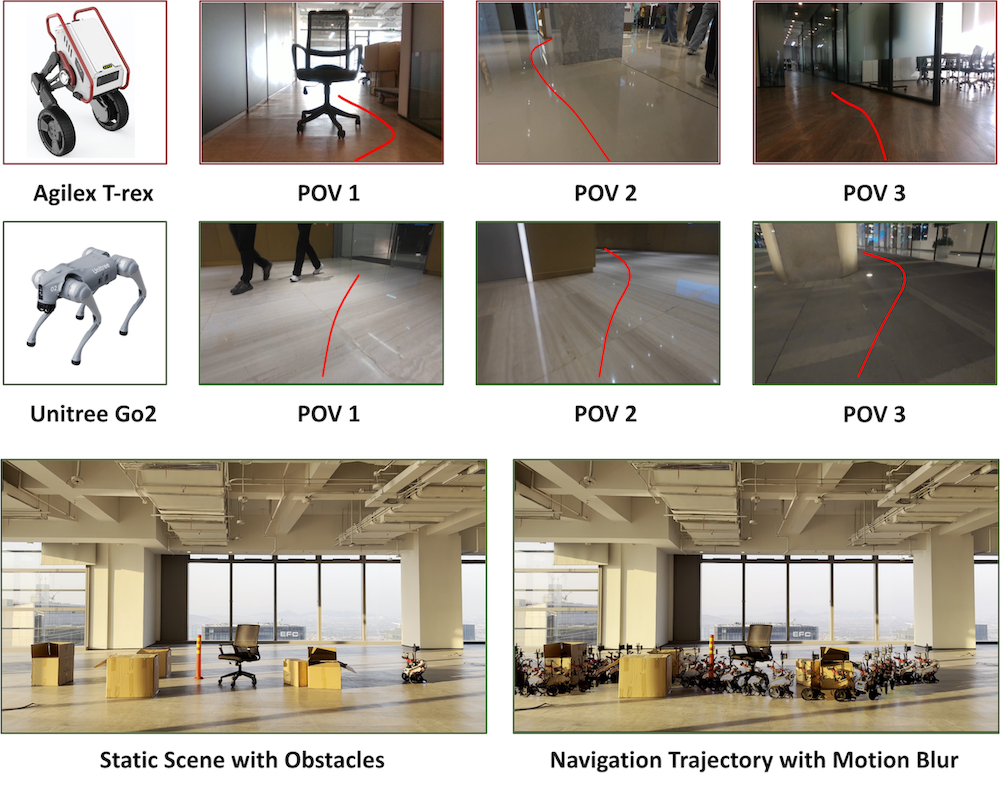} 
    \caption{
    Qualitative results. Top: Multi-platform POV. Bottom: 3rd-person static (L) and motion-blurred safe trajectories (R).
    }
    \label{fig:qualitative_results}
\vspace{-4mm}
\end{figure}

\subsubsection{Comparison with DAgger}

\begin{figure}
    \centering
    \includegraphics[width=0.9\linewidth]{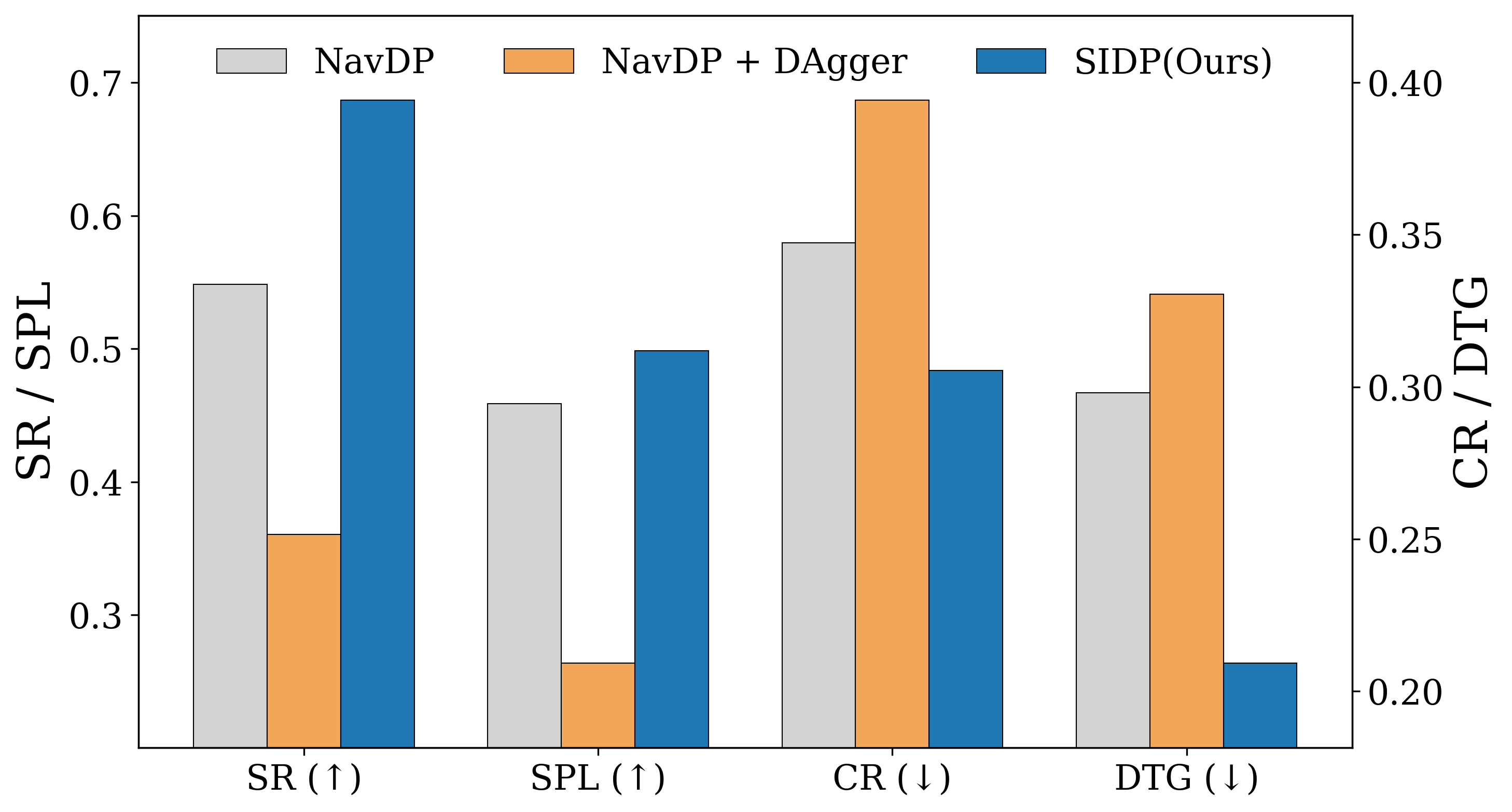}
    \caption{Performance comparison with DAgger. 
    }
    \label{fig:navdp_vs_dagger}
\vspace{-4mm}
\end{figure}

To address distribution shifts, we compare SIDP with Synthetic DAgger. Lacking an interactive simulator, we perturb states in InternData-N1 and query a privileged expert. As in \mbox{\cref{fig:navdp_vs_dagger}}, NavDP+DAgger unexpectedly underperforms vanilla NavDP due to: (1) Causal confusion: mimicking an expert with unobservable global states (e.g., maps) causes instability. (2) Catastrophic forgetting: aggregating out-of-distribution (OOD) recovery states distorts the nominal distribution. Conversely, SIDP generates corrective trajectories within its own representational space. Filtered by the reward-driven curriculum, SIDP mitigates OOD issues, ensuring distributional consistency and stable gradients.

\subsection{Real-world Deployment}

\cref{fig:qualitative_results} shows SIDP’s performance on \textit{Unitree Go2} and \textit{Agilex T-rex}. Stereo RGB images are processed via BANet~\cite{tsai2022banet} for depth-based reactive navigation, consistently yielding collision-free trajectories (\cref{fig:qualitative_results}, bottom right).

As in Tab {\ref{tab:real_world_sr_single}}, SIDP achieves 90.25\% mSR, outperforming NavDP by 7.0\%. Despite gait-induced oscillations on the quadruped, SIDP maintains a robust 88.00\% SR, confirming reliable navigation and obstacle avoidance across heterogeneous platforms.

\begin{table}[h!]
\centering
\small
\caption{Real-world Success Rate (SR, \%) Comparison.} 
\label{tab:real_world_sr_single}
\setlength{\tabcolsep}{6pt} 
\begin{tabular}{lccc}
\toprule
\textbf{Method} & \textbf{Agilex T-rex} & \textbf{Unitree Go2} & \textbf{mSR} ($\uparrow$) \\
\midrule
iPlanner     & 62.50 & 55.00 & 58.75 \\
ViPlanner    & 71.00 & 64.50 & 67.75 \\
NavDP        & 85.00 & 81.50 & 83.25 \\
\midrule
\textbf{SIDP (Ours)} & \textbf{92.50} & \textbf{88.00} & \textbf{90.25} \\
\bottomrule
\end{tabular}
\vspace{-6mm}
\end{table}

\section{Conclusion}

Self-Imitated Diffusion Policy (SIDP) integrates self-imitation with diffusion for robust visual navigation. By learning from self-generated experiences, SIDP bypasses dataset suboptimality and complex trajectory selection, minimizing overhead. Goal-agnostic exploration and reward-driven curricula enhance regularization and data utility. SIDP offers a scalable, high-performance solution for reactive path planning in complex environments.
Results show SIDP outperforms baselines in SR and SPL. Notably, SIDP achieves a 2.5$\times$ inference speedup on embedded platforms, confirming its suitability for real-time robotic deployment.
In conclusion, SIDP provides a robust and efficient solution for visual navigation, bridging the gap between generative models and practical robotic applications. In future work, the policy can be further improved from aspects of network architecture, input representation, and data composition. For instance, one promising direction is to explore more efficient generative paradigms such as flow matching within our framework.

\bibliographystyle{IEEEtran.bst}
\bibliography{refer}

\vfill

\end{document}